\documentclass[10pt, conference, compsocconf]{IEEEtran}

\usepackage{url}
\usepackage{verbatim}
\usepackage{epsfig}
\usepackage{ifthen}
\usepackage{graphics}
\usepackage{times}
\usepackage{amsfonts}
\usepackage{amsmath}
\usepackage{amssymb}
\usepackage{algorithmic,algorithm}

\newcommand{\bfx}{{\textbf{x}}}
\newcommand{\bfv}{{\textbf{v}}}
\newcommand{\bfu}{{\textbf{u}}}
\newcommand{\bfw}{{\textbf{w}}}
\newcommand{\bfy}{{\textbf{y}}}

\newcommand{\bfmu}{{\boldsymbol{\mu}}}
\newcommand{\bfpi}{{\boldsymbol{\pi}}}

\newcommand{\bfphi}{{\boldsymbol{\phi}}}
\newcommand{\bfvarphi}{{\boldsymbol{\varphi}}}

\begin{document}

\title{A novel transfer learning method based on common space mapping and weighted domain matching}

\author{
\IEEEauthorblockN{Ru-Ze Liang}
\IEEEauthorblockA{
King Abdullah University of Science\\
and Technology, Saudi Arabia\\
ruzeliang@outlook.com}
\and
\IEEEauthorblockN{Wei Xie}
\IEEEauthorblockA{
Vanderbilt University,\\
Nashville, TN 37235, United States\\
wei.xie@vanderbilt.edu}
\and
\IEEEauthorblockN{Weizhi Li}
\IEEEauthorblockA{
Suning Commerce R\&D Center USA, Inc\\
Palo Alto, CA 94304, United States\\
weizhili2014@gmail.com}
\and
\IEEEauthorblockN{Hongqi Wang}
\IEEEauthorblockA{
School of Management of Harbin\\
University of Science and Technology,\\
 Harbin 150000, China}
\and
\IEEEauthorblockN{Jim Jing-Yan Wang}
\IEEEauthorblockA{New York University Abu Dhabi,\\
United Arab Emirates}
\and
\IEEEauthorblockN{Lisa Taylor}
\IEEEauthorblockA{Michigan State University,\\
East Lansing, MI 48824, United States\\
lisataylor23857@yahoo.com}
}

\maketitle

\begin{abstract}
In this paper, we propose a novel learning framework for the problem of domain transfer learning. We map the data of two domains to one single common space, and learn a classifier in this common space. Then we adapt the common classifier to the two domains by adding two adaptive functions to it respectively. In the common space, the target domain data points are weighted and matched to the target domain in term of distributions. The weighting terms of source domain data points and the target domain classification responses are also regularized by the local reconstruction coefficients. The novel transfer learning framework is evaluated over some benchmark cross-domain data sets, and it outperforms the existing state-of-the-art transfer learning methods.
\end{abstract}


\IEEEpeerreviewmaketitle

\section{Introduction}
\label{sec:intro}

Transfer learning has been a hot topic in the machine learning community. It aims to solve the classifier learning problem of a target domain which has limited label information with week supervision information \cite{tan2016robust}, with the help of a source domain which has sufficient labels. The problem of using two domains for the problem of one domain is that their distributions are significantly different. A lot of works have been proposed to learn from two domains with different distributions for the classification problem in the target domain \cite{Chen20131284,Chu6619295,ma2014knowledge,Xiao201554,Li20141134,shao2015face}. However, the performance of these works are not satisfying. The shortages of these works paper are due to the ignorance of the label information of the target domain, the ignorance of the local connection of the data points of both source and target domain, or the ignorance of the the differences of the source domain data points for the learning problem of target domain.

\begin{figure}
  \centering
  \includegraphics[width=0.5\textwidth]{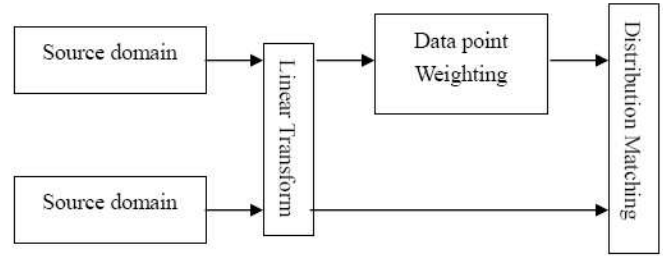}\\
  \caption{The distribution matching framework of our learning method.}\label{fig:matching}
\end{figure}

\begin{figure*}
  \centering
  \includegraphics[width=0.7\textwidth]{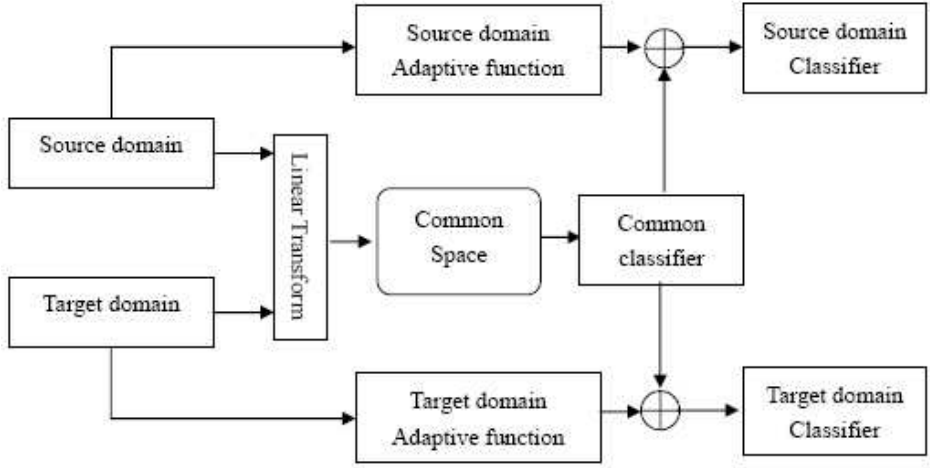}\\
  \caption{The learning framework of both source and target domain classifiers.}\label{fig:classifier}
\end{figure*}

In this paper, we propose a novel transfer learning problem to solve this problems. We map the data of two domains to one common space by linear transformation, and match the distribution of the two domains in this commons space. In these common subspaces, we match the distributions by using the weighting factors of the source domain data points. The distribution matching framework is shown in Figure. \ref{fig:matching}. We propose to minimize the classification errors of the data points of both the source and target domains to use the labels of the target domain. To do this, we learn a classifier in the common space by using the labels of data points of both domains, and then adapt the common classifier to the two domains by adding adaptive functions to the common classifier respectively. The learning framework of source and target domain classifiers is given in Figure. \ref{fig:classifier}. Moreover, we also propose to use local reconstruction information to regularize the learning of the weights of the source domain data points, and the classifier of the target domain. The learning problem is constructed by minimizing the objective function with regard to the parameters of the linear transformation matrix, the common classifier parameter and the adaptation parameters. We design an iterative learning algorithm to solve this problem.

The paper is organized as follows. In section \ref{sec:model}, we introduce the model and the learning method. In section \ref{sec:exp}, the proposed algorithm is evaluated over some benchmark data sets. In section \ref{sec:conclu}, the conclusion is given with the future works.

\section{Proposed transfer learning method}
\label{sec:model}

\subsection{Modeling}

We suppose the source domain training set is $\mathcal{S} = \{(\bfx_1^s, y_1^s), \cdots, (\bfx_{n_1}^s, y_{n_1}^s)\}$, where $\bfx_i^s\in \mathbb{R}^m$ is the feature vector of $m$ dimensions of the $i$-th data point, and $y_i^s\in \{+1,-1\}$ is its label. The target domain training set is $\mathcal{T} = \{(\bfx_1^t,y_1^t), \cdots, (\bfx_{n_3}^t,y_{n_3}^t), \bfx_{n_3+1}^t,\cdots, \bfx_{n_2}^t\}$, where $\bfx_j^t\in \mathbb{R}^m$ is the feature vector of the $j$-th data point, and $y_j^t\in \{+1,-1\}$ is its label. Only the the first $n_3$ target domain data points are labeled. We map the data of both domains to a common space by a transformation matrix $\Theta\in \mathbb{R}^{r\times m}$,

\begin{equation}
\label{equ:y}
\begin{aligned}
\bfy = \Theta \bfx.
\end{aligned}
\end{equation}
We present the distribution of the source domains in the common space as the weighted mean of the vectors of the data points,

\begin{equation}
\label{equ:mus}
\begin{aligned}
&\bfmu_s^\bfpi = \frac{1}{n_1} \sum_{i=1}^{n_1} \Theta \bfx_i^s \pi_i.
\end{aligned}
\end{equation}
where $\pi_i$ is the weighting factor of the $i$-th data point. We also present the distribution of the target domain as the mean of its data points in the common space,

\begin{equation}
\label{equ:mu}
\begin{aligned}
\bfmu_t = \frac{1}{n_2} \sum_{j=1}^{n_2} \Theta \bfx_j^t
\end{aligned}
\end{equation}
%
Naturally we hope the distributions of the two domains can be as close to each other as possible. So we propose to minimize the squared $\ell_2$ norm distance between them with regard to both $\Theta$ and $\pi$,

\begin{equation}
\label{equ:match}
\begin{aligned}
\min_{\Theta,\pi} ~~ & \frac{1}{2} \| \bfmu_s^\bfpi - \bfmu_t\|_2^2\\
& = \frac{1}{2} \left \| \frac{1}{n_1} \sum_{i=1}^{n_1} \Theta \bfx_i^s \pi_i -
\frac{1}{n_2} \sum_{j=1}^{n_2} \Theta \bfx_j^t \right \|_2^2.
\end{aligned}
\end{equation}

We design a linear classifier in the common space as in (\ref{equ:y}),

\begin{equation}
\label{equ:g}
\begin{aligned}
g(\bfx) = \bfw^\top \bfy = \bfw^\top \Theta \bfx,
\end{aligned}
\end{equation}
where $\bfw \in \mathbb{R}^r$ is the parameter vector of the common classifier $g$. Then we adapt it to two domains by adding adaptive functions to the common classifier, and obtain the source domain classifier $f$, and the target domain classifier $g$,

\begin{equation}
\label{equ:fs}
\begin{aligned}
&f(\bfx^s) = g(\bfx^s) +\Delta_s(\bfx^s) = \bfw^\top \Theta \bfx^s + \bfu^\top \bfx^s,~and\\
&h(\bfx^t) = g(\bfx^t) +\Delta_t(\bfx^t) = \bfw^\top \Theta \bfx^t + \bfv^\top \bfx^t,
\end{aligned}
\end{equation}
where $\Delta_s(\bfx^s) = \bfu^\top \bfx^s $ is the source domain adaptive function and $\bfu \in \mathbb{R}^m$ is its parameter vector of the adaptation function. where $\Delta_t(\bfx^t) = \bfv^\top \bfx^t$ is the target domain adaptive function, and $\bfv\in \mathbb{R}^m$ is its parameter vector. To measure the classification errors of the two classifiers over the training set, we use the popular hinge loss, and minimize it to learn the parameters,

\begin{equation}
\label{equ:loss}
\begin{aligned}
\min_{\Theta,\bfw,\bfu,\bfv,\bfpi} &\left \{
\sum_{i=1}^{n_1} \pi_i \max(0, 1-y_i^s f(\bfx^s_i))
\right.\\
&\left.
+ \sum_{j=1}^{n_3} \max(0, 1-y_i^t h(\bfx^t_j))
\right \}.
\end{aligned}
\end{equation}
In this classification error minimization problem, we also use the source domain data point weighting factors to weight the classification error terms.

We denote the neighborhood set of the $i$-th source data point as $\mathcal{N}_i^s$, and the reconstruction coefficients of $\mathcal{N}_i^s$ are solved by the following minimization problem,

\begin{equation}
\label{equ:reconstruction}
\begin{aligned}
\min_{\omega_{ik},k\in \mathcal{N}_i^s}
~&\left \|\bfx_i^s - \sum_{k\in \mathcal{N}_i^s} \omega_{ik}^s \bfx_k^s\right \|_2^2\\
s.t.~&\sum_{k\in \mathcal{N}_i^s} \omega_{ik}^s  =1,\omega_{ik}^s \geq 0,\forall~k\in \mathcal{N}_i^s
\end{aligned}
\end{equation}
where $\omega_{ik}^s ,k\in \mathcal{N}_i^s$ are the coefficients for reconstruction of $\bfx_i^s$ from the neighbors in $\mathcal{N}_i^s$. Then we use them to regularize the learning of the source domain weighting factors,

\begin{equation}
\label{equ:regularize1}
\begin{aligned}
\min_{\bfpi}
\sum_{i=1}^{n_1}  \left \|\pi_i - \sum_{k\in \mathcal{N}^s_i} \omega_{ik}^s \pi_k \right \|_2^2.
\end{aligned}
\end{equation}
Similarly we also have the neighborhood reconstruction coefficients for the target domain data set, and we use them to regularize the classification responses,

\begin{equation}
\label{equ:regularize2}
\begin{aligned}
\min_{\Theta,\bfw,\bfv}
\sum_{j=1}^{n_2}  \left \|h(\bfx^t_j) - \sum_{k'\in \mathcal{N}^t_j} \omega_{jk'}^t h(\bfx^t_{k'}) \right \|_2^2.
\end{aligned}
\end{equation}

The overall minimization problem for the transfer learning framework is the combination of problems of (\ref{equ:match}), (\ref{equ:loss}), (\ref{equ:regularize1}), (\ref{equ:regularize2}), and squared $\ell_2$ norms of classifier parameter vectors for over-fitting problems,

\begin{equation}
\label{equ:minim}
\begin{aligned}
\min_{\Theta,\bfw,\bfu,\bfv,\bfpi}~&\left \{
\sum_{i=1}^{n_1} \pi_i \max(0, 1-y_i^s f(\bfx^s_i))
\vphantom{\left.+\frac{C_3 }{2} \left \| \frac{1}{n_1} \sum_{i=1}^{n_1} \Theta \bfx_i^s \pi_i -
\frac{1}{n_2} \sum_{j=1}^{n_2} \Theta \bfx_j^t \right \|_2^2
\right\}}
\right.
\\
&
+ \sum_{j=1}^{n_3} \max(0, 1-y_i^t h(\bfx^t_j))\\
&+\frac{C_1}{2} \left ( \|\bfu\|_2^2 + \|\bfv\|_2^2 \right )\\
&+C_2 \left ( \sum_{i=1}^{n_1}  \left \|\pi_i - \sum_{k\in \mathcal{N}^s_i} \omega_{ik}^s \pi_k \right \|_2^2
\right.\\
&\left.+
\sum_{j=1}^{n_2}  \left \|h(\bfx^t_j) - \sum_{k'\in \mathcal{N}^t_j} \omega_{jk'}^t h(\bfx^t_{k'}) \right \|_2^2
 \right )\\
&\left.+\frac{C_3 }{2} \left \| \frac{1}{n_1} \sum_{i=1}^{n_1} \Theta \bfx_i^s \pi_i -
\frac{1}{n_2} \sum_{j=1}^{n_2} \Theta \bfx_j^t \right \|_2^2
\right\}\\
s.t.~&\Theta \Theta^\top = I_r,\\
&\bf0 \leq\bfpi \leq \delta\bf1, ~and~\bfpi^\top \bf1 = n_1.
\end{aligned}
\end{equation}
In this minimization problem, we impose $\Theta$ to be orthogonal, impose a lower bound and a upper bound  for $\bfpi$, and an additional constraint to $\bfpi$, so that the summation of all the elements of $\bfpi$ is $n_1$.

\subsection{Optimizatoin}

We rewrite the source domain and target domain classifiers as a linear function of the input feature vectors,

\begin{equation}
\label{equ:fs1}
\begin{aligned}
&f(\bfx^s) = \bfphi^\top \bfx^s, ~where~
\bfphi=\Theta^\top\bfw + \bfu,~and\\
&h(\bfx^t) = \bfvarphi^\top \bfx^t, ~where~
\bfvarphi= \Theta^\top\bfw+\bfv.
\end{aligned}
\end{equation}
%
Then we have the following minimization problem,

\begin{equation}
\label{equ:minim}
\begin{aligned}
\min_{\Theta,\bfw,\bfphi,\bfvarphi,\bfpi}~&\left \{
\sum_{i=1}^{n_1} \pi_i \max(0, 1-y_i^s \bfphi^\top \bfx^s_i))
\vphantom{\left.+\frac{C_3 }{2} \left \| \frac{1}{n_1} \sum_{i=1}^{n_1} \Theta \bfx_i^s \pi_i -
\frac{1}{n_2} \sum_{j=1}^{n_2} \Theta \bfx_j^t \right \|_2^2
\right\}}
\right.
\\
&
+ \sum_{j=1}^{n_3} \max(0, 1-y_i^t \bfvarphi^\top\bfx^t_j)\\
&+\frac{C_1}{2} \left ( \|\bfphi - \Theta^\top\bfw\|_2^2 + \|\bfvarphi - \Theta^\top\bfw\|_2^2 \right )\\
&+C_2 \left ( \sum_{i=1}^{n_1}  \left \|\pi_i - \sum_{k\in \mathcal{N}^s_i} \omega_{ik}^s \pi_k \right \|_2^2
\right.\\
&\left.+
\sum_{j=1}^{n_2}  \left \|\bfvarphi^\top \bfx^t_j - \sum_{k'\in \mathcal{N}^t_j} \omega_{jk'}^t \bfvarphi^\top \bfx^t_{k'} \right \|_2^2
 \right )\\
&\left.+\frac{C_3 }{2} \left \| \frac{1}{n_1} \sum_{i=1}^{n_1} \Theta \bfx_i^s \pi_i -
\frac{1}{n_2} \sum_{j=1}^{n_2} \Theta \bfx_j^t \right \|_2^2
\right\}\\
s.t.~&\Theta \Theta^\top = I_r,\\
&\bf0 \leq\bfpi \leq \delta\bf1, ~and~\bfpi^\top \bf1 = n_1.
\end{aligned}
\end{equation}
To solve this problem, we use the iterative optimization method to update the variables one by one.

\subsubsection{Solving $\bfw$ and $\Theta$}

We first solve $\bfw$ by setting the derivative of objective with regard to $\bfw$ to zero, and we have

\begin{equation}
\label{equ:woptimal}
\begin{aligned}
\bfw = \frac{1}{2}\Theta (\bfphi + \bfvarphi).
\end{aligned}
\end{equation}
Then we substitute it to (\ref{equ:minim}), and consider the optimization of $\Theta$, we have

\begin{equation}
\label{equ:minTheta1}
\begin{aligned}
\min_{\Theta}~
&
Tr\left[\Theta \left (
\vphantom{\left ( \frac{1}{n_1} \sum_{i=1}^{n_1}  \bfx_i^s \pi_i -
\frac{1}{n_2} \sum_{j=1}^{n_2}  \bfx_j^t \right ) ^\top}
-\frac{C_1}{4}  (\bfphi+\bfvarphi)(\bfphi+\bfvarphi)^\top
+\frac{C_3}{2}
\left ( \frac{1}{n_1} \sum_{i=1}^{n_1}  \bfx_i^s \pi_i
\right.
\right.
\right.
\\
&
\left.\left.\left.
-\frac{1}{n_2} \sum_{j=1}^{n_2}  \bfx_j^t \right )
\left ( \frac{1}{n_1} \sum_{i=1}^{n_1}  \bfx_i^s \pi_i -
\frac{1}{n_2} \sum_{j=1}^{n_2}  \bfx_j^t \right ) ^\top
\right ) \Theta^\top \right]\\
s.t.~&
\Theta \Theta^\top = I_r,
\end{aligned}
\end{equation}
This problem can be easily solve by the eigen-decomposition method.

\subsubsection{Updating $\bfphi$ and $\bfvarphi$}

To update both $\bfphi$ and $\bfvarphi$, we consider the following minimization problem,

\begin{equation}
\label{equ:minimuv}
\begin{aligned}
\min_{\bfphi,\bfvarphi}~&  \left \{ \mathcal{Q}(\bfphi,\bfvarphi) =
\sum_{i=1}^{n_1} \max(0, 1-y_i^s \bfphi^\top \bfx^s_i)) \pi_i \right.\\
&+ \sum_{j=1}^{n_3} \max(0, 1-y_i^t \bfvarphi^\top\bfx^t_j)\\
&+\frac{C_1}{2} \left ( \|\bfphi - \Theta^\top\bfw\|_2^2 + \|\bfvarphi - \Theta^\top\bfw\|_2^2 \right )\\
&\left .+C_2
\sum_{j=1}^{n_2}  \left \|\bfvarphi^\top \bfx^t_j - \sum_{k'\in \mathcal{N}^t_j} \omega_{jk'}^t \bfvarphi^\top \bfx^t_{k'} \right \|_2^2 \right \}.
\end{aligned}
\end{equation}
To solve this problem, we use the sub-gradient algorithm to update $\bfphi$ and $\bfvarphi$,

\begin{equation}
\label{equ:updateuv}
\begin{aligned}
\bfphi\leftarrow \bfphi - \rho \nabla\mathcal{Q}_{\bfphi}, ~and~
\bfvarphi\leftarrow \bfvarphi - \rho \nabla\mathcal{Q}_{\bfvarphi},
\end{aligned}
\end{equation}
The sub-gradient functions of $\mathcal{Q}$ with regard to $\bfphi$ and $\bfvarphi$ are

\begin{equation}
\label{equ:suggra}
\begin{aligned}
&\nabla\mathcal{Q}_{\bfphi}=
- \sum_{i=1}^{n_1} \alpha_i y_i^s \bfx^s_i \pi_i +C_1 (\bfphi - \Theta^\top\bfw),\\
&where~\alpha_i=1,~if~(1-y_i^s \bfphi^\top \bfx^s_i) \geq 0,~and~0~otherwise.\\
&\nabla\mathcal{Q}_{\bfvarphi}=
-\sum_{j=1}^{n_3} \beta_j y_i^t \bfx^t_j +C_1 (\bfvarphi - \Theta^\top\bfw)\\
&+2 C_2
\sum_{j=1}^{n_2}   \left ( \bfx^t_j - \sum_{k'\in \mathcal{N}^t_j} \omega_{jk'}^t \bfx^t_{k'} \right )
\left ( \bfx^t_j - \sum_{k'\in \mathcal{N}^t_j} \omega_{jk'}^t \bfx^t_{k'} \right )^\top \bfvarphi,\\
&where~\beta_j = 1,~if~(1-y_i^t \bfvarphi^\top\bfx^t_j)\geq 0,~and~0~otherwise.
\end{aligned}
\end{equation}

\subsubsection{Updating $\bfpi$}

To solve $\bfpi$, we have the following minimization problem,

\begin{equation}
\label{equ:minimpi}
\begin{aligned}
\min_{\bfpi}~
&\left\{\sum_{i=1}^{n_1} \max(0, 1-y_i^s \bfphi^\top \bfx^s_i)) \pi_i\right.\\
&+C_2 \sum_{i=1}^{n_1}  \left \|\pi_i - \sum_{k\in \mathcal{N}^s_i} \omega_{ik}^s \pi_k \right \|_2^2\\
&\left.+\frac{C_3 }{2} \left \| \frac{1}{n_1} \sum_{i=1}^{n_1} \Theta \bfx_i^s \pi_i -
\frac{1}{n_2} \sum_{j=1}^{n_2} \Theta \bfx_j^t \right \|_2^2\right\},\\
s.t.~&\bf0 \leq\bfpi \leq \delta\bf1, ~and~\bfpi^\top \bf1 = n_1.
\end{aligned}
\end{equation}
This problem is a linear constrained quadratic programming problem, and we solve it by using the active set algorithm.

\section{Experiments}
\label{sec:exp}

\subsection{Data Sets}

In the experiments, we use three benchmark data sets. Which are the 20-Newsgroup corpus data set, the Amazon review data set, and the Spam email data set.  {20-Newsgroup corpus data set} is a data set of newspaper documents. It contains documents of 20 classes. The classes are organized in a hierarchical structure. For a class, it usually have two or more sub-classes. For example, in the class of \emph{car}, there are two sub-classes, which are \emph{motorcycle} and \emph{auto}. To split this data set to source domain and target domain, for one class, we keep one sub-class in the source domain, while put the other sub-class to the target domain. We follow the splitting of source and target domain of NG14 data set of \cite{Chen20131284}. In this data set, there are 6 classes, and for each class, one sub-class is in the source domain, and another sub-class is in the target domain. For each domain, the number of data points is 2,400. The bag-of-word features of each document are used as original features. {Amazon review data set} is a data set of reviews of products. It contains reviews of three types of products, which are books, DVD and Music. The reviews belongs to two classes, which are positive and negative. We treat the review of books as source domain, and that of DVD as target domain. For each domain, we have 2,000 positive reviews and 2,000 reviews. Again, we use the bag-of-words features as the features of reviews. {Spam email data set} is a set of emails of different individuals. In this data set, there are emails of three different individuals' inboxes, and we treat each individual as a domain. In each individual's inbox, there are 2,500 emails, and the emails are classified to two different classes, which are normal email and spam email. we also randomly choose one individual as a source domain, and another one as a target domain.

\subsection{Results}

In the experiments, we use the $10$-fold cross validation. For each data set, we use each domain as a target domain in turns, and randomly choose anther domain as a source domain. The classification accuracies of the compared methods over three benchmark data sets are reported in Table \ref{tab:acc}. The proposed method outperforms all the compared methods over three benchmark data sets. In the experiments over the 20-Newsgroup data set, the proposed method outperforms the other methods significantly.

\begin{table}
\centering
\caption{Classification accuracy of compared methods over benchmark data sets.}
\label{tab:acc}
\begin{tabular}{|l||r|r|r|r|r|}
\hline
Methods & 20-Newsgroup & Amazon & Spam \\\hline
\hline
Proposed                        &0.6210&0.7812&0.8641\\\hline
Chen et al. \cite{Chen20131284} &0.5815&0.7621&0.8514\\\hline
Chu et al. \cite{Chu6619295}    &0.5471&0.7642&0.8354\\\hline
Ma et al. \cite{ma2014knowledge}&0.5164&0.7255&0.8012\\\hline
Xiao and Guo \cite{Xiao201554}  &0.5236&0.7462&0.8294\\\hline
Li et al. \cite{Li20141134}     &0.5615&0.7134&0.8122\\\hline
\end{tabular}
\end{table}

\section{Conclusions}
\label{sec:conclu}

In this paper, we proposed a novel transfer learning method. Instead of learning a common representation and classifier directly for both source and target domains, we proposed to learn common space and classifier, and then adapt it to source and target domains. We proposed to weight the source domain data points in the subspaces to match the distributions of the two domains, and to regularize the weighting factors of the source domain data points and the classification responses of the target domain data points by the local reconstruction coefficients.The minimization problem of our method is based on these features, and we solve it by an iterative algorithm. Experiments show its advantages over some other methods.
In the future, we will extend the proposed algorithm to various applications, such as computational mechanic \cite{wang2014computational,zhou2014biomarker,liu2013structure,peng2015modeling,xu2016mechanical,zhou2016mechanical}, 
multimedia\cite{liang2016semi,liang2016optimizing,manifoldlearning,generalized,wang2014effective,lin2016multi,liu2015supervised,wang2015multiple,wang2015supervised}, medical imaging \cite{king2015surgical,li2015outlier,thatcher2016multispectral,li2015burn,squiers2016multispectral,chenhui2015spectral,hu2016matched,chen2014integrating,su2015virtual}, bioinformatics \cite{xie2014securema,li2016supporting,cai2011optimization,cai2013modeling}, material science \cite{zhuge2009random,kang2009comparison,zhang2015selective}, high-performance computing \cite{wang2007gradual,luo2014federated,Meng2015,Meng2014,Meng2013}, malicious websites detection \cite{xu2014evasion,xu2013cross,xu2014adaptive,xu2012push}, biometrics \cite{wang2014leveraging,wang2015new,wang2013new,wang2013gender}, etc. We will also consider using some other models to represent and construction the classifier, such as Bayesian network \cite{fan2014tightening,fan2014finding,fan2015improved}.


\begin{thebibliography}{10}
\providecommand{\url}[1]{#1}
\csname url@rmstyle\endcsname
\providecommand{\newblock}{\relax}
\providecommand{\bibinfo}[2]{#2}
\providecommand\BIBentrySTDinterwordspacing{\spaceskip=0pt\relax}
\providecommand\BIBentryALTinterwordstretchfactor{4}
\providecommand\BIBentryALTinterwordspacing{\spaceskip=\fontdimen2\font plus
\BIBentryALTinterwordstretchfactor\fontdimen3\font minus
  \fontdimen4\font\relax}
\providecommand\BIBforeignlanguage[2]{{%
\expandafter\ifx\csname l@#1\endcsname\relax
\typeout{** WARNING: IEEEtran.bst: No hyphenation pattern has been}%
\typeout{** loaded for the language `#1'. Using the pattern for}%
\typeout{** the default language instead.}%
\else
\language=\csname l@#1\endcsname
\fi
#2}}

\bibitem{tan2016robust}
M.~Tan, Z.~Hu, B.~Wang, J.~Zhao, and Y.~Wang, ``Robust object recognition via
  weakly supervised metric and template learning,'' \emph{Neurocomputing}, vol.
  181, pp. 96--107, 2016.

\bibitem{Chen20131284}
B.~Chen, W.~Lam, I.~Tsang, and T.-L. Wong, ``Discovering low-rank shared
  concept space for adapting text mining models,'' \emph{IEEE Transactions on
  Pattern Analysis and Machine Intelligence}, vol.~35, no.~6, pp. 1284--1297,
  2013.

\bibitem{Chu6619295}
W.~S. Chu, F.~D.~L. Torre, and J.~F. Cohn, ``Selective transfer machine for
  personalized facial action unit detection,'' in \emph{Computer Vision and
  Pattern Recognition (CVPR), 2013 IEEE Conference on}, 2013, pp. 3515--3522.

\bibitem{ma2014knowledge}
Z.~Ma, Y.~Yang, N.~Sebe, and A.~G. Hauptmann, ``Knowledge adaptation with
  partiallyshared features for event detectionusing few exemplars,''
  \emph{Pattern Analysis and Machine Intelligence, IEEE Transactions on},
  vol.~36, no.~9, pp. 1789--1802, 2014.

\bibitem{Xiao201554}
M.~Xiao and Y.~Guo, ``Feature space independent semi-supervised domain
  adaptation via kernel matching,'' \emph{IEEE Transactions on Pattern Analysis
  and Machine Intelligence}, vol.~37, no.~1, pp. 54--66, 2015.

\bibitem{Li20141134}
W.~Li, L.~Duan, D.~Xu, and I.~Tsang, ``Learning with augmented features for
  supervised and semi-supervised heterogeneous domain adaptation,'' \emph{IEEE
  Transactions on Pattern Analysis and Machine Intelligence}, vol.~36, no.~6,
  pp. 1134--1148, 2014.

\bibitem{shao2015face}
H.~Shao, S.~Chen, J.-y. Zhao, W.-c. Cui, and T.-s. Yu, ``Face recognition based
  on subset selection via metric learning on manifold,'' \emph{Frontiers of
  Information Technology \& Electronic Engineering}, vol.~16, pp. 1046--1058,
  2015.

\bibitem{wang2014computational}
S.~Wang, Y.~Zhou, J.~Tan, J.~Xu, J.~Yang, and Y.~Liu, ``Computational modeling
  of magnetic nanoparticle targeting to stent surface under high gradient
  field,'' \emph{Computational mechanics}, vol.~53, no.~3, pp. 403--412, 2014.

\bibitem{zhou2014biomarker}
Y.~Zhou, W.~Hu, B.~Peng, and Y.~Liu, ``Biomarker binding on an
  antibody-functionalized biosensor surface: the influence of surface
  properties, electric field, and coating density,'' \emph{The Journal of
  Physical Chemistry C}, vol. 118, no.~26, pp. 14\,586--14\,594, 2014.

\bibitem{liu2013structure}
Y.~Liu, J.~Yang, Y.~Zhou, and J.~Hu, ``Structure design of vascular stents,''
  \emph{Multiscale simulations and mechanics of biological materials}, pp.
  301--317, 2013.

\bibitem{peng2015modeling}
B.~Peng, Y.~Liu, Y.~Zhou, L.~Yang, G.~Zhang, and Y.~Liu, ``Modeling
  nanoparticle targeting to a vascular surface in shear flow through diffusive
  particle dynamics,'' \emph{Nanoscale research letters}, vol.~10, no.~1, pp.
  1--9, 2015.

\bibitem{xu2016mechanical}
J.~Xu, J.~Yang, N.~Huang, C.~Uhl, Y.~Zhou, and Y.~Liu, ``Mechanical response of
  cardiovascular stents under vascular dynamic bending,'' \emph{Biomedical
  engineering online}, vol.~15, no.~1, p.~1, 2016.

\bibitem{zhou2016mechanical}
Y.~Zhou, S.~Sohrabi, J.~Tan, and Y.~Liu, ``Mechanical properties of nanoworm
  assembled by dna and nanoparticle conjugates,'' \emph{Journal of Nanoscience
  and Nanotechnology}, vol.~16, no.~6, pp. 5447--5456, 2016.

\bibitem{liang2016semi}
R.-Z. Liang, W.~Xie, W.~Li, X.~Du, J.~J.-Y. Wang, and J.~Wang,
  ``Semi-supervised structured output prediction by local linear regression and
  sub-gradient descent,'' \emph{arXiv preprint arXiv:1606.02279}, 2016.

\bibitem{liang2016optimizing}
R.-Z. Liang, L.~Shi, H.~Wang, J.~Meng, J.~J.-Y. Wang, Q.~Sun, and Y.~Gu,
  ``Optimizing top precision performance measure of content-based image
  retrieval by learning similarity function,'' in \emph{Pattern Recognition
  (ICPR), 2016 23st International Conference on}.\hskip 1em plus 0.5em minus
  0.4em\relax IEEE, 2016.

\bibitem{manifoldlearning}
M.~Ding and G.~Fan, ``Multilayer joint gait-pose manifolds for human gait
  motion modeling,'' \emph{IEEE Transactions on Cybernetics}, vol.~45, no.~11,
  pp. 2413--2424, 2015.

\bibitem{generalized}
------, ``Articulated and generalized gaussian kernel correlation for human
  pose estimation,'' \emph{IEEE Transactions on Image Processing}, vol.~25,
  no.~2, pp. 776--789, 2016.

\bibitem{wang2014effective}
H.~Wang and J.~Wang, ``An effective image representation method using kernel
  classification,'' in \emph{Tools with Artificial Intelligence (ICTAI), 2014
  IEEE 26th International Conference on}.\hskip 1em plus 0.5em minus
  0.4em\relax IEEE, 2014, pp. 853--858.

\bibitem{lin2016multi}
F.~Lin, J.~Wang, N.~Zhang, J.~Xiahou, and N.~McDonald, ``Multi-kernel learning
  for multivariate performance measures optimization,'' \emph{Neural Computing
  and Applications}, pp. 1--13, 2016.

\bibitem{liu2015supervised}
X.~Liu, J.~Wang, M.~Yin, B.~Edwards, and P.~Xu, ``Supervised learning of sparse
  context reconstruction coefficients for data representation and
  classification,'' \emph{Neural Computing and Applications}, pp. 1--9, 2015.

\bibitem{wang2015multiple}
J.~Wang, H.~Wang, Y.~Zhou, and N.~McDonald, ``Multiple kernel multivariate
  performance learning using cutting plane algorithm,'' in \emph{Systems, Man,
  and Cybernetics (SMC), 2015 IEEE International Conference on}.\hskip 1em plus
  0.5em minus 0.4em\relax IEEE, 2015, pp. 1870--1875.

\bibitem{wang2015supervised}
J.~Wang, Y.~Zhou, K.~Duan, J.~J.-Y. Wang, and H.~Bensmail, ``Supervised
  cross-modal factor analysis for multiple modal data classification,'' in
  \emph{Systems, Man, and Cybernetics (SMC), 2015 IEEE International Conference
  on}.\hskip 1em plus 0.5em minus 0.4em\relax IEEE, 2015, pp. 1882--1888.

\bibitem{king2015surgical}
D.~R. King, W.~Li, J.~J. Squiers, R.~Mohan, E.~Sellke, W.~Mo, X.~Zhang, W.~Fan,
  J.~M. DiMaio, and J.~E. Thatcher, ``Surgical wound debridement sequentially
  characterized in a porcine burn model with multispectral imaging,''
  \emph{Burns}, vol.~41, no.~7, pp. 1478--1487, 2015.

\bibitem{li2015outlier}
W.~Li, W.~Mo, X.~Zhang, J.~J. Squiers, Y.~Lu, E.~W. Sellke, W.~Fan, J.~M.
  DiMaio, and J.~E. Thatcher, ``Outlier detection and removal improves accuracy
  of machine learning approach to multispectral burn diagnostic imaging,''
  \emph{Journal of biomedical optics}, vol.~20, no.~12, pp. 121\,305--121\,305,
  2015.

\bibitem{thatcher2016multispectral}
J.~E. Thatcher, W.~Li, Y.~Rodriguez-Vaqueiro, J.~J. Squiers, W.~Mo, Y.~Lu,
  K.~D. Plant, E.~Sellke, D.~R. King, W.~Fan, \emph{et~al.}, ``Multispectral
  and photoplethysmography optical imaging techniques identify important tissue
  characteristics in an animal model of tangential burn excision,''
  \emph{Journal of Burn Care \& Research}, vol.~37, no.~1, pp. 38--52, 2016.

\bibitem{li2015burn}
W.~Li, W.~Mo, X.~Zhang, Y.~Lu, J.~J. Squiers, E.~W. Sellke, W.~Fan, J.~M.
  DiMaio, and J.~E. Thatcher, ``Burn injury diagnostic imaging device's
  accuracy improved by outlier detection and removal,'' in \emph{SPIE Defense+
  Security}.\hskip 1em plus 0.5em minus 0.4em\relax International Society for
  Optics and Photonics, 2015, pp. 947\,206--947\,206.

\bibitem{squiers2016multispectral}
J.~J. Squiers, W.~Li, D.~R. King, W.~Mo, X.~Zhang, Y.~Lu, E.~W. Sellke, W.~Fan,
  J.~M. DiMaio, and J.~E. Thatcher, ``Multispectral imaging burn wound tissue
  classification system: a comparison of test accuracies of several common
  machine learning algorithms,'' 2016.

\bibitem{chenhui2015spectral}
J.~S. K. A. J. G. E. F. Y. M. L. Q.~L. Chenhui~Hu, Lin~Cheng, ``A spectral
  graph regression model for learning brain connectivity of alzheimer's
  disease,'' \emph{PLOS ONE}, 2015.

\bibitem{hu2016matched}
C.~Hu, J.~Sepulcre, K.~A. Johnson, G.~E. Fakhri, Y.~M. Lu, and Q.~Li, ``Matched
  signal detection on graphs: Theory and application to brain imaging data
  classification,'' \emph{NeuroImage}, vol. 125, pp. 587--600, 2016.

\bibitem{chen2014integrating}
J.-J. Chen and W.-C. Su, ``Integrating iso/iwa1 practices with an automatic
  patient image quantity examination approach to achieve the patient safety
  goal in a nuclear medicine department,'' in \emph{Internet of Things
  (iThings), 2014 IEEE International Conference on, and Green Computing and
  Communications (GreenCom), IEEE and Cyber, Physical and Social Computing
  (CPSCom), IEEE}.\hskip 1em plus 0.5em minus 0.4em\relax IEEE, 2014, pp.
  332--335.

\bibitem{su2015virtual}
W.-C. Su, S.-C. Yeh, S.-H. Lee, and H.-C. Huang, ``A virtual reality lower-back
  pain rehabilitation approach: System design and user acceptance analysis,''
  in \emph{International Conference on Universal Access in Human-Computer
  Interaction}.\hskip 1em plus 0.5em minus 0.4em\relax Springer, 2015, pp.
  374--382.

\bibitem{xie2014securema}
W.~Xie, M.~Kantarcioglu, W.~S. Bush, D.~Crawford, J.~C. Denny, R.~Heatherly,
  and B.~A. Malin, ``Securema: protecting participant privacy in genetic
  association meta-analysis,'' \emph{Bioinformatics}, p. btu561, 2014.

\bibitem{li2016supporting}
W.~Li, H.~Liu, P.~Yang, and W.~Xie, ``Supporting regularized logistic
  regression privately and efficiently,'' \emph{PloS one}, vol.~11, no.~6, p.
  e0156479, 2016.

\bibitem{cai2011optimization}
W.~Cai, X.~Zhou, and X.~Cui, ``Optimization of a gpu implementation of
  multi-dimensional rf pulse design algorithm,'' in \emph{Bioinformatics and
  Biomedical Engineering,(iCBBE) 2011 5th International Conference on}.\hskip
  1em plus 0.5em minus 0.4em\relax IEEE, 2011, pp. 1--4.

\bibitem{cai2013modeling}
W.~Cai and L.~L. Gouveia, ``Modeling and simulation of maximum power point
  tracker in ptolemy,'' \emph{Journal of Clean Energy Technologies}, vol.~1,
  no.~1, 2013.

\bibitem{zhuge2009random}
J.~Zhuge, L.~Zhang, R.~Wang, R.~Huang, D.-W. Kim, D.~Park, and Y.~Wang,
  ``Random telegraph signal noise in gate-all-around silicon nanowire
  transistors featuring coulomb-blockade characteristics,'' \emph{Applied
  Physics Letters}, vol.~94, no.~8, p. 083503, 2009.

\bibitem{kang2009comparison}
Z.~Kang, L.~Zhang, R.~Wang, and R.~Huang, ``A comparison study of silicon
  nanowire transistor with schottky-barrier source/drain and doped
  source/drain,'' in \emph{VLSI Technology, Systems, and Applications, 2009.
  VLSI-TSA'09. International Symposium on}.\hskip 1em plus 0.5em minus
  0.4em\relax IEEE, 2009, pp. 133--134.

\bibitem{zhang2015selective}
L.~Zhang, H.~Li, Y.~Guo, K.~Tang, J.~Woicik, J.~Robertson, and P.~C. McIntyre,
  ``Selective passivation of geo2/ge interface defects in atomic layer
  deposited high-k mos structures,'' \emph{ACS applied materials \&
  interfaces}, vol.~7, no.~37, pp. 20\,499--20\,506, 2015.

\bibitem{wang2007gradual}
C.-Y. Wang, D.-Y. Peng, L.~Xu, and X.-S. Yi, ``Gradual gray-watermark embedding
  algorithm in the wavelet domain [j],'' \emph{Journal of Computer
  Applications}, vol.~6, p. 025, 2007.

\bibitem{luo2014federated}
W.~Luo, L.~Xu, Z.~Zhan, Q.~Zheng, and S.~Xu, ``Federated cloud security
  architecture for secure and agile clouds,'' in \emph{High Performance Cloud
  Auditing and Applications}.\hskip 1em plus 0.5em minus 0.4em\relax Springer
  New York, 2014, pp. 169--188.

\bibitem{Meng2015}
L.~Meng and J.~Johnson, ``High performance implementation of the inverse tft,''
  in \emph{Proceedings of the 2015 International Workshop on Parallel Symbolic
  Computation}, ser. PASCO '15, 2015, pp. 87--94.

\bibitem{Meng2014}
------, ``High performance implementation of the tft,'' in \emph{Proceedings of
  the 39th International Symposium on Symbolic and Algebraic Computation}, ser.
  ISSAC '14, 2014, pp. 328--334.

\bibitem{Meng2013}
------, \emph{Automatic Parallel Library Generation for General-Size Modular
  FFT Algorithms}, 2013, pp. 243--256.

\bibitem{xu2014evasion}
L.~Xu, Z.~Zhan, S.~Xu, and K.~Ye, ``An evasion and counter-evasion study in
  malicious websites detection,'' in \emph{Communications and Network Security
  (CNS), 2014 IEEE Conference on}.\hskip 1em plus 0.5em minus 0.4em\relax IEEE,
  2014, pp. 265--273.

\bibitem{xu2013cross}
------, ``Cross-layer detection of malicious websites,'' in \emph{Proceedings
  of the third ACM conference on Data and application security and
  privacy}.\hskip 1em plus 0.5em minus 0.4em\relax ACM, 2013, pp. 141--152.

\bibitem{xu2014adaptive}
S.~Xu, W.~Lu, L.~Xu, and Z.~Zhan, ``Adaptive epidemic dynamics in networks:
  Thresholds and control,'' \emph{ACM Transactions on Autonomous and Adaptive
  Systems (TAAS)}, vol.~8, no.~4, p.~19, 2014.

\bibitem{xu2012push}
S.~Xu, W.~Lu, and L.~Xu, ``Push-and pull-based epidemic spreading in networks:
  Thresholds and deeper insights,'' \emph{ACM Transactions on Autonomous and
  Adaptive Systems (TAAS)}, vol.~7, no.~3, p.~32, 2012.

\bibitem{wang2014leveraging}
X.~Wang and C.~Kambhamettu, ``Leveraging appearance and geometry for kinship
  verification,'' in \emph{2014 IEEE International Conference on Image
  Processing (ICIP)}.\hskip 1em plus 0.5em minus 0.4em\relax IEEE, 2014, pp.
  5017--5021.

\bibitem{wang2015new}
------, ``A new approach for face recognition under makeup changes,'' in
  \emph{2015 IEEE Global Conference on Signal and Information Processing
  (GlobalSIP)}.\hskip 1em plus 0.5em minus 0.4em\relax IEEE, 2015, pp.
  423--427.

\bibitem{wang2013new}
X.~Wang, V.~Ly, G.~Guo, and C.~Kambhamettu, ``A new approach for 2d-3d
  heterogeneous face recognition,'' in \emph{Multimedia (ISM), 2013 IEEE
  International Symposium on}.\hskip 1em plus 0.5em minus 0.4em\relax IEEE,
  2013, pp. 301--304.

\bibitem{wang2013gender}
X.~Wang and C.~Kambhamettu, ``Gender classification of depth images based on
  shape and texture analysis,'' in \emph{Global Conference on Signal and
  Information Processing (GlobalSIP), 2013 IEEE}.\hskip 1em plus 0.5em minus
  0.4em\relax IEEE, 2013, pp. 1077--1080.

\bibitem{fan2014tightening}
X.~Fan, C.~Yuan, and B.~Malone, ``Tightening bounds for bayesian network
  structure learning,'' in \emph{Proceedings of the 28th AAAI Conference on
  Artificial Intelligence (AAAI-2014)}, 2014.

\bibitem{fan2014finding}
X.~Fan, B.~Malone, and C.~Yuan, ``Finding optimal bayesian network structures
  with constraints learned from data,'' in \emph{Proceed. of the 30th Conf. on
  Uncertainty in Artificial Intelligence (UAI-2014)}, 2014.

\bibitem{fan2015improved}
X.~Fan and C.~Yuan, ``An improved lower bound for bayesian network structure
  learning,'' in \emph{Proceedings of the 29th AAAI Conference on Artificial
  Intelligence (AAAI-2015)}, 2015.

\end{thebibliography}

\end{document}